\documentclass{article}

\usepackage[nonatbib, preprint]{neurips_2025}

\usepackage[utf8]{inputenc} 
\usepackage[T1]{fontenc}    
\usepackage{hyperref}       
\usepackage{url}            
\usepackage{booktabs}       
\usepackage{amsfonts}       
\usepackage{nicefrac}       
\usepackage{microtype}      
\usepackage{xcolor}         

\usepackage[numbers]{natbib}

\usepackage{styles/macros}

\usepackage{amsmath,amsfonts,bm}

\def\eqref#1{equation~\ref{#1}}

\def\1{\bm{1}}

\DeclareMathAlphabet{\mathsfit}{\encodingdefault}{\sfdefault}{m}{sl}
\SetMathAlphabet{\mathsfit}{bold}{\encodingdefault}{\sfdefault}{bx}{n}

\DeclareMathOperator*{\argmax}{arg\,max}

\usepackage{amsmath}
\usepackage{amssymb}
\usepackage{mathtools}
\usepackage{amsthm}

\usepackage{tcolorbox}
\usepackage{caption}
\usepackage{subcaption}

\usepackage{listings}
\usepackage[compatibility=false]{caption}
\usepackage{booktabs}
\definecolor{lightblue}{RGB}{95, 158, 160} 

\usepackage[normalem]{ulem}
\usepackage{paralist}
\usepackage{mathtools}
\usepackage{xspace}
\usepackage{tikz}
\usepackage{amsmath}
\usetikzlibrary{shapes,arrows}
\usetikzlibrary{positioning}
\usetikzlibrary{arrows.meta}
\usepackage{pgfplots}
\usepackage{subcaption}
\usepackage{booktabs}
\usepackage{tcolorbox}
\usepackage{wrapfig}
\usepackage{hyperref}
\usepackage{tabularx}
\usepackage{booktabs}
\usepackage{array}

\newcommand{\bx}{\mathbf{x}}

\newcommand{\bz}{\mathbf{z}}

\newcommand{\ourmethod}{\texttt{BOLT}}
\newcommand{\ourmethodfullname}{Bayesian Optimization with LLM Transfer (\texttt{BOLT})}

\usepackage[inline]{enumitem}
\usepackage[ruled,vlined]{algorithm2e}
\usepackage{adjustbox}

\usepackage[capitalize,noabbrev]{cleveref}

\title{Large Scale Multi-Task Bayesian Optimization with Large Language Models}

\author{%
Yimeng Zeng$^{1}$ \quad Natalie Maus$^{1}$ \quad Haydn Thomas Jones$^1$ \quad Jeffrey Tao$^1$ \\
\textbf{Fangping Wan}$^2$  \quad
\textbf{Marcelo Der Torossian Torres}$^2$ \\ \textbf{Cesar de la Fuente-Nunez}$^2$ \quad \textbf{Ryan Marcus}$^1$ \quad \textbf{Osbert Bastani}$^1$ \quad \textbf{Jacob R. Gardner}$^1$\\
$^1$Computer and Information Science, UPenn \quad $^2$Perelman School of Medicine, UPenn\\
\texttt{yimengz@seas.upenn.edu}
}

\begin{document}

\maketitle

\begin{abstract}
In multi-task Bayesian optimization, the goal is to leverage experience from optimizing existing tasks to improve the efficiency of optimizing new ones. While approaches using multi-task Gaussian processes or deep kernel transfer exist, the performance improvement is marginal when scaling beyond a moderate number of tasks. We introduce a novel approach leveraging large language models (LLMs) to learn from, and improve upon, previous optimization trajectories, scaling to approximately 1500 distinct tasks. Specifically, we propose a feedback loop in which an LLM is fine-tuned on the high quality solutions to specific tasks found by Bayesian optimization (BO). This LLM is then used to generate initialization points for future BO searches for new tasks. The trajectories of these new searches provide additional training data for fine-tuning the LLM, completing the loop. We evaluate our method on two distinct domains: database query optimization and antimicrobial peptide design. Results demonstrate that our approach creates a positive feedback loop, where the LLM's generated initializations gradually improve, leading to better optimization performance. As this feedback loop continues, we find that the LLM is eventually able to generate solutions to new tasks \emph{in just a few shots} that are better than the solutions produced by ``from scratch'' by Bayesian optimization while simultaneously requiring significantly fewer oracle calls.
\end{abstract}
\section{Introduction}

\vspace{-1.5ex}
Multi-task optimization seeks to use related, previously solved tasks to accelerate the optimization of new ones. Multi-task optimization appears naturally in a variety of domains where similar problems are encountered repeatedly, such as hyperparameter optimization, material science,  database query optimization, and drug design. Formally, suppose we have tasks $\{1, 2, \ldots, T\}$, each associated with its own objective function $f_t(\mathbf{x})$. For each task $t \in \{1, 2, \ldots, T\}$, we seek to find some $\mathbf{x}^*_t$ such that 
\begin{equation}
    \mathbf{x}_t^* = \underset{\mathbf{x}\in \mathcal{X}}{\arg\min} f_t(\mathbf{x}).
\end{equation}
We focus on the setting where, for each task, we have collected a dataset $D_{t}$ of observations, and we wish to leverage this data when optimizing unseen test tasks.

Multi-task Bayesian optimization (BO) methods have predominantly used Gaussian processes (GPs) and/or a variety of shared weight neural network feature extractors to jointly learn correlations across tasks \citep{swersky_multi-task_2013, perrone_scalable_2018, dkt, hakhamaneshi_jumbo_2022}. A standard approach involves placing a multi-output GP over the input-task space, decomposing the kernel as an input kernel $k(\mathbf{x}, \mathbf{x}^\prime)$ and a task kernel $k(t, t^\prime)$. Despite their effectiveness, many of these methods --- with the notable exception of recent work such as~\citet{wang_pre-trained_2024} --- tend to saturate in performance after tens of training tasks and do not extract additional performance improvement on new tasks when given hundreds or thousands of related tasks. 

We propose \ourmethodfullname{}, a straightforward approach to multi-task BO that departs from the framework of building related task information into the BO surrogate model. Instead, as BO completes optimization for training tasks, we fine-tune a large language model (LLM) to, given a task description or context $C[f_{t}]$, generate solutions for that optimization problem that we can use as strong initialization for BO.  

This approach creates a self-reinforcing feedback loop: BO generates high-quality solutions that we can leverage to fine-tune the LLM; the fine-tuned LLM, in turn, produces better initializations that improve BO performance. Over time, the LLM learns to directly generate solutions that are highly competitive, enabling top-$k$-samples from the LLM (requiring just a few oracle calls) to outperform full ``from scratch'' BO runs (requiring a large number of oracle calls). This iterative improvement enables \ourmethod{} to scale and still extract value from thousands of tasks. We validate \ourmethod{} on two diverse and challenging domains where many related tasks are available. 

\textbf{Application 1: Database query plan optimization.} In database query plan optimization, the goal is to find query plans for SQL queries that minimize runtime. Efficient query planning is critical for database performance~\cite{howgood}, and traditional optimizers, such as PostgreSQL, rely on hand-crafted heuristics and cost models~\cite{systemr}. 
We show that an LLM fine-tuned using \ourmethod{} can generate superior plans, sometimes outperforming PostgreSQL's own optimizer in a single-shot setting. 

\textbf{Application 2: Antimicrobial peptide (AMP) design.} Antimicrobial peptides are small peptides that kill (pathogenic) bacteria. We use a software package, APEX \citep{apex}, which can predict the minimum inhibitory concentration (MIC) of peptides against various pathogens. 
In this setting, each task is a distinct \emph{extinct} peptide used as a template, and the goal is to enhance its antimicrobial activity through a limited number of edits.
We demonstrate that our approach continues to improve as more tasks are introduced and, in the few-shot setting, can eventually generate peptides that outperform those found through full ``from-scratch'' BO runs. 

Our experimental results highlight two key benefits of \ourmethod{}. First, unlike many existing multi-task BO methods, LLM-generated initialization continues to improve performance as the number of tasks grows, avoiding the saturation observed in GP-based methods. Impressively, by the end of training, the LLM-generated initializations alone nearly outperform competing approaches. Second, a sufficiently fine-tuned LLM can eventually produce solutions in a few-shot setting that rival or surpass full ``from scratch'' BO runs. Notably, in the database query optimization setting, the LLM becomes a better one-shot optimizer than PostgreSQL's built-in query planner for certain query types. 

\textbf{Contributions} 
\begin{enumerate}[itemsep=0ex,leftmargin=*]
\vspace{-2ex}
\item We propose \ourmethod{}, a scalable and \textit{simple} alternative to traditional multi-task BO, leveraging LLMs to generate strong initial solutions for new tasks. \ourmethod{} leverages a combination of high quality optimized solutions produced by BO and self augmentation for fine-tuning.
\item We validate \ourmethod{} through experiments on two challenging domains: database query optimization and antimicrobial peptide design. Experimental results demonstrate that initializations generated by \ourmethod{} continue to improve performance as the number of tasks grows. 
\item We provide empirical results demonstrating that, after sufficient fine-tuning, the LLM alone can eventually match or even outperform full ``from scratch'' Bayesian optimization runs with significantly fewer oracle calls.

\end{enumerate}

\section{Background}

\vspace{-1.5ex}
\paragraph{Bayesian optimization (BO).} 
Bayesian Optimization (BO) \cite{movckus1975bayesian, SnoekBO} is an iterative approach to optimize black-box functions in a sample-efficient manner. 
On each step of the optimization, a supervised probabilistic \textit{surrogate model} (usually a Gaussian Process (GP) \cite{rasmussen2003gaussian}) is conditioned on all data collected so far. 
Then, the surrogate model's predictive posterior distribution $p(y \mid \bx, D)$ is used to decide what data point(s) should be evaluated next, typically by maximizing some \textit{acquisition function}, defined with respect to $p(y \mid \bx, D)$, which guides the exploration-exploitation trade off. 
Finally, selected points are evaluated on the black-box function and added to the dataset. 
This iterative process continues until the evaluation budget is reached. 

\vspace{-1.5ex}
\paragraph{Structured optimization via latent space BO.} 
BO has recently been applied to optimizing structured search spaces, such as molecular and amino acid sequences, by leveraging latent space Bayesian optimization. This approach incorporates a variational autoencoder (VAE) to map structured inputs into a continuous latent space, where BO is performed \cite{eissman2018bayesian, Weighted_Retraining, Huawei, siivola2021good, lambo, lolbo}. Structured inputs $\bx$ (e.g., amino acid sequences) are mapped to continuous latent representations $\bz$ by the VAE encoder $\Phi(\bx)$. This creates a transformed continuous (latent) representation of the structured search space where BO can be directly applied. The corresponding latent candidate points are then decoded by the VAE decoder, $\Gamma(\bz)$, to reconstruct structured outputs for evaluation.
For large combinatorial structured search spaces, such as the space of organic molecules or the space of all peptide amino acid sequences, the latent space of the VAE is typically high-dimensional (on the order of several hundred dimensions) in order to represent the large structured space effectively.

\vspace{-1.5ex}
\paragraph{Optimizing antimicrobial peptides.} 
In antimicrobial peptide design, we seek peptides (sequences of amino acids) that minimize the MIC (minimum inhibitory concentration, measured in $\mu$ mol $\text{L}^{-1}$) for some target bacterial pathogen. MIC is a measure of the concentration of the peptide required to inhibit growth of the target bacterial pathogen \citep{kowalska2021minimum}. 
A key challenge in antimicrobial peptide design is that many modern bacterial pathogens have developed resistance to modern antibiotics. To solve this challenge, \citet{apex} propose designing new peptides with high sequence similarity to template peptides mined from extinct organisms. The template peptides themselves do not typically achieve sufficiently low MIC for target bacteria pathogens. However, since these template peptides have not been encountered in nature for thousands of years, modern antimicrobial resistant bacteria have not evolved resistance to them. It follows that new peptides are more likely to evade antibiotic resistance if they are designed to be similar to the extinct template sequences. We employ this strategy, optimizing antimicrobial peptides with a minimum threshold sequence similarity to the extinct template peptides from \citet{apex}. 
We also employ latent space BO to optimize over the structured space of amino acid sequences.

\vspace{-1.5ex}
\paragraph{Optimizing database query plans.}
Query optimization in data management systems involves translating a declarative SQL query into an execution plan that efficiently retrieves the correct results~\cite{volcano}. This problem has been extensively investigated in the field of data management~\cite{job2}, as the difference in execution time between an optimal and a poorly chosen query plan can be several orders of magnitude~\cite{howgood}.
Since individual query plans are composed of discrete characteristics (e.g. join order trees), the search space of possible query plans is structured and combinatorial. 
We therefore employ latent space BO. 
We use the string representation for query plans proposed by~\citet{sigmod-arxiv} to pre-train a VAE model that maps the structured space of query plans to a continuous latent space where BO can be applied. 

\vspace{-1.5ex}
\paragraph{Database query plan optimization with right-censored observations.}
In database query optimization, our black-box objective function measures the execution latency of the query plan.
``Good'' and ``bad'' query plans can have latencies differing by multiple orders of magnitude~\cite{howgood}. 
This can lead to the majority of optimization runtime being taken up by evaluating a small number of poorly performing plans. 
A natural solution to this problem is to \textit{time out} objective function evaluations after they have reached some threshold latency $\tau$, resulting in \textit{right-censored} observations.
A right-censored observation is an observation at data point $\bx$ where we observe only that $y \geq \tau$ for some chosen timeout threshold $\tau$, rather than observing the typical noisy objective value $y$. 
Prior work has been done to extend Bayesian optimization methods to the setting of right-censored observations. 
\citet{DBLP:journals/corr/HutterHL13, eggensperger2020neural} extended Bayesian optimization methods to the setting of right-censored observations by introducing an EM-like algorithm to impute the values of censored observations.
\citet{eggensperger2020neural} expanded on this, defining a single surrogate model capable of being conditioned on the combination of censored and uncensored data gathered. 

\citet{sigmod-arxiv} extend this to the setting of approximate GP surrogate models. 
Since we focus on tasks that involve large function evaluation budgets, we employ \citet{sigmod-arxiv}'s proposed method of modeling censored data with approximate GPs.  

\section{\ourmethodfullname{}}
\vspace{-1.5ex}

We propose \ourmethodfullname{}, an iterative framework for using large language models (LLMs) to improve Bayesian optimization (BO) performance across a family of related tasks. We are given a set of $T$ training tasks defined by objective functions $f_{1}(\bx),...,f_{T}(\bx)$. We additionally assume that, for each objective function we have a \textit{context} or \textit{task description} $C[f_{t}]$ that can be a natural language or other input description that differentiates $f_{t}$ from any other task in the application domain. For example, this might be the text of a SQL query we are trying to optimize.

For each \textit{training} task, we assume we have optimized the objective with some BO procedure, resulting in the optimization trajectories $\{\mathcal{D}^\star_{t}\}_{t=1}^T$, with each $\mathcal{D}^\star_{t}$ containing the top-$K$ observations from the trajectory for the $t^\text{th}$ task. Our goal is to leverage this training data to learn an LLM-based ``initialization policy'' $\pi$ that, when presented with new related tasks $\left\{f_{T+1}(\bx), C[f_{T+1}(\bx)]\right\}$, proposes a high-quality set of candidate solutions for BO to further refine. 

These two procedures --- (1) using BO to collect high-quality data for training tasks, and (2) using the LLM to initialize BO for new tasks --- can be used as an ``outer-loop''/``inner-loop'' approach to solving a large number of related tasks sequentially, where the LLM is periodically updated as more optimization runs complete.

Because the LLM and BO only interact through generating initialization and generating fine-tuning data respectively, our approach here is relatively agnostic to the specific underlying implementation of BO used to optimize each task. This enables the straightforward use of the full range of recent BO advances on high-dimensional, constrained, and other optimization settings.

\begin{minipage}{0.48\linewidth}
\begin{algorithm}[H]
\SetAlgoVlined
\DontPrintSemicolon
\SetKwInOut{Require}{Require}
\SetKwInOut{Ensure}{Ensure}
\caption{Inner Loop: LLM-Initialized Bayesian Optimization}
\small
\Require{Task $t$, context $C[f_t]$, LLM $\pi_n$, budget $B$, batch $b$}
\Ensure{Optimized solutions $X^*_t$}
$X_{\text{init}} \gets \pi_n(C[f_t])$ \tcp*{LLM proposes candidates}
Evaluate $y_{\text{init}} \gets f_t(X_{\text{init}})$\;
$\mathcal{D} \gets (X_{\text{init}}, y_{\text{init}})$\;
Initialize $\mathcal{GP}(X_{\text{init}}, y_{\text{init}})$\;
\For{step $i = 1$ \KwTo $\lfloor B/b \rfloor$}{
  $X_{\text{next}} \gets \argmax_{\alpha}(x; \mathcal{GP})$ \tcp*{Acquire}
  $y_{\text{next}} \gets f_t(X_{\text{next}})$\;
  $\mathcal{D} \gets (X \cup X_{\text{next}}, y \cup y_{\text{next}})$\;
  Update $\mathcal{GP}$ with new observations\;
}
Return $X^*_t \gets \text{top-}K(X)$ \tcp*{Best solutions}
\label{alg:inner}
\end{algorithm}
\end{minipage}
\hfill
\begin{minipage}{0.48\linewidth}
\begin{algorithm}[H]
\SetAlgoVlined
\DontPrintSemicolon
\SetKwInOut{Require}{Require}
\SetKwInOut{Ensure}{Ensure}
\caption{Outer Loop: LLM Fine-Tuning via BO Trajectories}
\small
\Require{Dataset $\mathcal{D}_0 = \{(C[f_t], x_i, y_i)\}$, LLM $\pi_0$, iterations $T$}
\Ensure{Fine-tuned LLM $\pi_T$}
Initialize $\mathcal{D} \gets \mathcal{D}_0$, $\pi \gets \pi_0$\;
\For{iteration $k = 1$ \KwTo $T$}{
  \ForEach{\textbf{task} $t$ in batch}{
    $X^*_{t} \gets$ \textsc{InnerLoop}($t$, $\pi_k$, $B$, $b$) \tcp*{Run BO}
    \vspace{1pt}
    $\mathcal{D} \gets \mathcal{D} \cup \{(C[f_t], x, y) \,|\, x \in X^*_{t}\}$\;
    \vspace{1pt}
    \tcp*{Augment with top solutions}
  }
  \vspace{1pt}
  Fine-tune $\pi_{k}$ on augmented dataset $\mathcal{D}$\;
  \vspace{1pt}
  Update model parameters via instruction prompting\;
}
Return $\pi_T$ \tcp*{Final fine-tuned LLM}
\label{alg:outer}
\end{algorithm}
\end{minipage}

\vspace{-1.5ex}
\paragraph{Initializing \ourmethod{}.} At initialization for a workload of tasks, we have only an un-tuned LLM \ourmethod{}-$0$ that is generally useless for the task setting because it is unaware of even the specific format for candidate suggestions. For the first iteration, we solve $T$ optimization tasks with a single-task BO routine where we initialize BO using some standard initialization procedure. We run optimization on each of the $T$ initial tasks, and extract the optimization trajectories $\{\mathcal{D}^{*}_{i}\}_{i=1}^T$ from each run.

\vspace{-1.5ex}
\paragraph{LLM fine-tuning.}
The LLM fine-tuning process employs supervised learning using OpenAI's \texttt{GPT-4-mini-0718} model through their API. From the optimization trajectories $\{\mathcal{D}^{*}_{i}\}_{i=1}^T$, we extract the top-$K$ observations from each of the $T$ runs completed so far. We use these observations along with the task contexts $\left\{C[f_{t}]\right\}_{t=1}^{T}$ to construct a fine-tuning dataset $\mathcal{D_{\textrm{ft}}}$. Each training instance contains: 
\begin{enumerate}[leftmargin=*]
\item A system prompt shared across all tasks in the workload/problem domain, which specifies the objective (e.g., generating efficient join orderings).
\item A user prompt with the task-specific context $C[f_t]$ (e.g., the SQL query requiring optimization).
\item A response prompt containing the high-performing solution $x$ discovered through BO.
\end{enumerate}

We fine-tune using OpenAI's standard fine-tuning API \cite{openai2024gpt4technicalreport}. Specifically, we format our data into the required JSONL format (i.e., prompt-solution pairs) and then upload it via the fine-tuning API to initiate training. The model is trained to minimize the negative log-likelihood of the solution tokens $\textbf{x}$ given the task context $C$:
\begin{equation}
\mathcal{L} = -\sum_{i=1}^{|\textbf{x}|} \log \pi(x_i | C, \textbf{x}_{<i})
\end{equation}

We note that our approach leverages full model fine-tuning rather than extensive and/or manual prompt engineering~\citep{softprompt, Li2021PrefixTuningOC}. This allows the model to learn the task requirements through the context-solution pairs in $\mathcal{D_{\textrm{ft}}}$, rather than explicit instructions. However, for scenarios requiring few-shot learning on untrained models, more careful prompt engineering may be beneficial.

This fine-tuning process produces an updated model that encodes the knowledge from $\mathcal{D_{\textrm{ft}}}$. In our experiments, we will refer to an LLM trained on $T$ tasks in this way as \ourmethod{}-$T$.

\vspace{-1.5ex}
\paragraph{LLM fine-tuning frequency.}
The number of tasks $T$ that we collect at initialization time and during each round of the \ourmethod{} ``outer-loop'' represents a non-trivial trade-off due to the computational cost of both running BO and the cost of fine-tuning the LLM. Fine-tuning the LLM more frequently results in both additional computational and monetary costs, but allows subsequent BO runs to complete more efficiently (with fewer black-box function evaluations). In this paper, we erred on the side of lower monetary cost in exchange for additional cost in black-box function evaluations. Specifically, we fine-tuned an LLM 4 times for the query plan optimization task and 7 times for the antimicrobial peptide design task as shown in \cref{fig:scatter-plots}

\vspace{-1.5ex}
\paragraph{Using the LLM for multi-task BO.}
Once we have a fine‐tuned model, \ourmethod{}-$T$, we can leverage the fine-tuned LLM's capabilities to generate higher-quality initialization points for subsequent optimization tasks. For a set of $n$ new tasks $\{t_i\}_{i=T+1}^{T+n}$, we sample from \ourmethod{}-$T$ to generate the same number of initialization points used by the baseline ``from scratch'' approach. The sampling prompt maintains the same structure as the training prompt without the assistant response. The \ourmethod{}-$T$ generated solutions are refined with a standard BO routine, and the top‐$K$ performing solutions for each task $t$—along with their contexts $C[f_t]$—are incorporated into the training set for the next round of fine‐tuning.

\textbf{Closed-loop feedback.} This iterative process creates a positive feedback loop between BO and the LLM. After completing each iteration of BO, we augment the LLM fine-tuning dataset with the newly collected high-scoring solutions and their corresponding contexts. Re-training the LLM on this expanded dataset allows it to internalize additional insights over the tasks and produce more effective initialization for the next batch of tasks. Through multiple iterations, the LLM learns to propose increasingly high-quality starting points, thereby jump-starting subsequent BO runs and reducing the time needed to find near‐optimal solutions. Ultimately, this positive feedback loop continually refines the model's understanding of what a good solution should look like under each context, improving its performance across the entire family of tasks.

\begin{wrapfigure}{r}{0.51\textwidth}
\begin{minipage}{\linewidth}
\begin{algorithm}[H]
\SetAlgoVlined
\DontPrintSemicolon
\SetKwInOut{Require}{Require}
\SetKwInOut{Ensure}{Ensure}
\caption{Self-augmentation for LLM Finetuning}
\small
\Require{Tasks $\mathcal{T}$, LLM $\pi_\theta$, iterations $T$, criteria $\mathcal{C}$}
\Ensure{Fine-tuned LLM $\pi_{\theta + T}$}
Sample $\mathcal{D} \gets \mathcal{D}_0$, $\pi \gets \pi_0$\;
\For{iteration $k = 1$ \textbf{to} $T$}{
  \ForEach{\textbf{task} $t \in \mathcal{T}$}{
    $X_{init} \sim \pi_{\theta + k}(t)$ \tcp*{Generate samples}
    $X^\star_{init} \gets SelectBest(X_{init}, C)$ \;
    \tcp*{Select best samples}
  }
  $\mathcal{D} \gets \mathcal{D} \cup \{(C[f_t], x, y) \,|\, x \in X^*_{init}\}$\;
  \tcp*{Augment dataset}
  Fine-tune $\pi_{\theta + k}$ on $\mathcal{D}$\;
}
Return $\pi_T$ \tcp*{Final fine-tuned LLM}
\label{alg:boot}
\end{algorithm}
\end{minipage}
\end{wrapfigure}

\vspace{-1.5ex}
\paragraph{Self-Augmentation.}
As the fine-tuned LLM enhances few-shot generation with more optimization data, it is worth exploring whether the costly sequential BO processes can be minimized. Thus, we explore ``self-improvement'' methods to refine the LLM policy without the expense of additional optimization runs (\cref{alg:boot}). Specifically, once an LLM has been fine-tuned using some of the tasks we set aside for training, we prompt it to generate additional solutions for \textit{all} available tasks in that problem setting. We then score these solutions using the problem's oracle and fine-tune the LLM again directly with this labeled self-generated data, in a manner similar to self-play in reinforcement learning or self-instruction in LLM training \citep{code_self_playhaluptzok2022, shypula2024code}. By filtering and fine-tuning on its own best outputs, the LLM can iteratively teach itself how to propose better solutions.

\section{Experiments}
\label{sec:experiments}

\vspace{-1.5ex}
We evaluate \ourmethod{} on two distinct problem domains, each with a large number of related tasks. For both domains, problem definitions and solutions can be represented as strings. This allows \ourmethod{} to operate both in sequence space, where the LLM learns from optimization trajectories, and in latent space, where BO makes additional progress using LLM-sampled initializations.

\vspace{-1.5ex}
\paragraph{Implementation details.} For the inner optimization loop, we implement a constrained version of the \texttt{LOL-BO} algorithm \citep{lolbo} using BoTorch and GPyTorch \citep{botorch, gpytorch}. Code to reproduce all results in the paper is available on GitHub: \url{placeholder}. For query optimization, we use an acquisition batch size of 1 with a budget of 4,000 oracle calls, while for peptide design, we employ a larger acquisition batch size of 50 with a budget of 200,000 oracle calls.

The outer loop \ourmethod{}-$T$ models uses instruction prompting \citep{mishra_reframing_2021, longpre2023flan} to guide the LLM in producing optimized sequences. \cref{fig:db_train_prompt} shows the prompt used to prompt \gptfourmini for efficient query plans. After each optimization iteration, we augment the training set with the highest-scoring sequences from the optimization trajectory and fine-tune \gptfourmini on this expanded dataset. When fine-tuning the LLM for the query plan optimization task, we use OpenAI's automatic batch size selection option. For the peptide design task, we found that using the automatic batch size option did not provide a similar boost in performance, and we use a constant batch size of $10$. For both tasks, we fine-tuned the LLM for $2$ epochs and used the default OpenAI LR multiplier hyperparameter of $1.8$. 
To ensure the solutions always have the correct syntax, we filter out characters that do not correspond to strings of integers or valid amino acids for the respective tasks. \cref{fig:peptide_train_prompt} provides additional details on the fine-tuning process and prompts used for the peptide task. 

\vspace{-1.5ex}
\paragraph{Database query plan optimization.} Database query plan optimization focuses on finding query plans (including join orderings and their operators) with low execution time for a given query. We take a subset of $2933$ queries from the Cardinality Estimation Benchmark introduced by \citet{ceb}, keeping $99$ queries for validation. 
Following \citet{sigmod-arxiv}, we perform BO over query plans by encoding join orders and operators as integer lists, which are then mapped to a 64-dimensional continuous latent space using the pre-trained query plan VAE from \citet{sigmod-arxiv}.
For the pretraining dataset, we randomly generate $1,169,890$ query plans generated based on the database schema, separated into 80/10/10 splits. For the initial ``from scratch'' runs with no LLM, we initialize with the set of 50 query plans used by BAO~\citep{bao}, an ML-powered query optimizer we use as a baseline, that produces reasonable but non-optimal plans. Subsequent runs use 50 LLM-sampled query plans per query as initialization points. All points are sampled using a temperature parameter of 0.7 unless otherwise specified.
The ``task description context" $C$ used to fine-tune the LLM for this task is the full SQL query string.

\vspace{-1.5ex}
\paragraph{Antimicrobial peptide design.} For the peptide design application, we are given a library of 1000 extinct, weakly antimicrobial seed peptides $S = \left\{s_{1},...,s_{L}\right\}$. A task in this setting is to take a particular seed peptide $s_{i}$ and make modifications to it to minimize the minimum inhibitory concentration (MIC) against \textit{A. Baumannii} ATCC $19606$, measured in $\mu \text{mol}/L$. We created a library of $L=1000$ extinct peptides and held out the last $100$ as validation. We ensure edited peptides maintain a minimum $75\%$ similarity to the seed peptide, defined by $1 - \frac{d(S, S')}{\text{len}(S)}$, where $d$ is the Levenshtein distance between them. All of the validation peptides are at least $25\%$ different from any other peptide in the library. Although the seed peptides don't achieve low MICs, the hope is that bacteria are less likely to have developed resistance to their variations as they come from extinct species \citep{apex}. We assess MICs with the APEX model \citet{apex} and utilize a VAE trained on 4.5 million amino acid sequences \citet{apexgo} to map peptides into a $256$-dimensional latent space. Initial optimization uses $1000$ randomly mutated sequences with a similarity constraint of $75\%$ to the seed. Subsequent runs utilize $1000$ LLM sampled peptides. All points are sampled using a temperature parameter of 1.0 unless otherwise specified. We use the seed amino acid sequence as the ``task description context'' $C$ for LLM fine-tuning.

\vspace{-1.5ex}
\paragraph{Baselines.} 
We compare \ourmethod{} against a range of baseline approaches. First, we compare to ``from scratch,'' single task \texttt{LOL-BO} which we will refer to as \texttt{STBO}, which operates without prior task knowledge. Second, we compare to a common strategy for multi task BO---see, e.g., \citet{dkt, hakhamaneshi_jumbo_2022, perrone_scalable_2018}, where a shared GP is trained on all tasks through a neural network feature extractor using the optimization trajectories from training tasks. This shared GP is then check-pointed and used on test tasks. Several papers have found success with variations of this approach. ABLR \cite{perrone_scalable_2018} uses independent Bayesian linear regression heads per task on the shared feature extractor, while FSBO \citep{wistuba2021few} uses an adaptation of DKT \citep{dkt}. In this paper, for the final supervised model, we use the same \texttt{PPGPR} model \citep{ppgpr} as the BO inner-loop in our method as we require scalability but find this results in better performance than Bayesian linear regression or random Fourier feature models \citep{rahimi_random_2007}. 

We also compare against methods that utilize an ensemble of Gaussian process experts, POGPE and SGPE \citep{pogpe}. POGPE utilizes an ensemble approach with one expert from each previous task, while SGPE extends this by adding an additional expert trained on data for the current task only with higher weighting. We evaluate several configurations of these methods (with 5, 10, and 20 experts) to identify optimal performance. Additionally, we evaluate two transformer-based methods: Optformer \citep{optformer} and LLAMBO \citep{llambo}. Optformer employs a fine-tuned transformer model (in our implementation, GPT-4o-mini) for hyperparameter optimization, while LLAMBO uses out-of-the-box LLMs (also GPT-4o-mini in our case, compared to GPT-3.5 in the original paper) for both surrogate modeling and acquisition function optimization. Due to context length limitations with these transformer-based methods, we maintain a sliding window of the last 100 oracle calls for both approaches. For LLAMBO, we impose a budget limit of 10 million input tokens per experiment to manage computational costs. Additional details on all baselines can be found in \cref{appendix:baseline_info}.

\vspace{-1.5ex}
\subsection{Optimization results}
\vspace{-1.5ex}

\begin{figure*}[t]
\centering
\includegraphics[width=\linewidth]{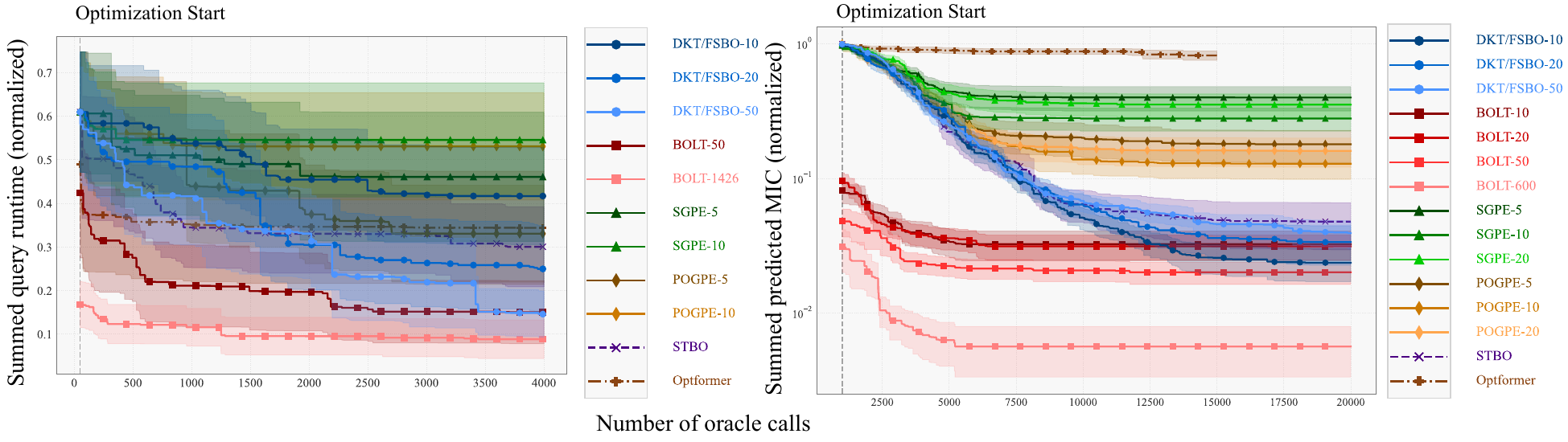}
\caption{Bayesian optimization performance on \textbf{(Left)} query plan optimization and \textbf{(Right)} antimicrobial peptide design. In both settings, \ourmethod{} outperforms or matches baselines with just initialization data before optimization begins.}
\label{fig:oracle_calls}
\vspace{-3ex}
\end{figure*}

In \cref{fig:oracle_calls}, we demonstrate that initializing BO with \ourmethod{} significantly improves optimization efficiency across both domains. On the query optimization task (left), while DKT/FSBO makes improvements over STBO, the gains appear to plateau after only 20 tasks. In contrast, \ourmethod{} successfully scales to over 1400 tasks and converges to higher quality solutions faster. On the peptide design task (right), \ourmethod{} shows similarly strong performance, while DKT/FSBO struggles to take advantage of the data collected for separate templates. 

The GP expert methods show mixed results. For the database task, POGPE-5 and SGPE-5 demonstrate better performance than their variations with more experts, while for the peptide task, POGPE-10 and SGPE-10 yield the best results compared to 5/20 experts. Consistent with findings from \citet{pogpe}, POGPE generally outperforms SGPE across both domains. However, both ensemble approaches are consistently outperformed by \ourmethod{}, and even fall behind STBO and DKT/FSBO in several cases. We did not run POGPE or SGPE with a larger numbers of experts as both methods scale poorly, requiring updating of a number of GPs proportional to the number of tasks.

\begin{figure*}[t]
\centering
\includegraphics[width=\linewidth]{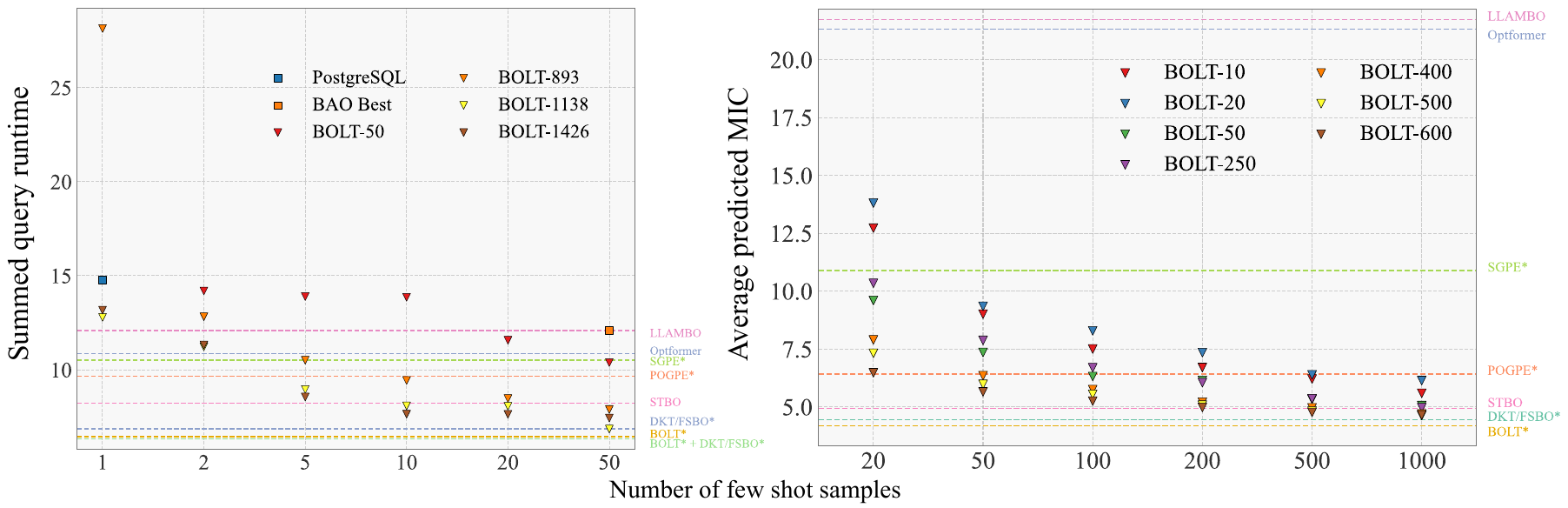}
\caption{Evaluating \ourmethod{} in the few shot setting and comparing to full optimization runs in both problem settings (\textbf{Left:} query plan optimization; \textbf{Right:} peptide design). In each plot, we show objective values accumulated across all validation tasks for various methods. Scatter points illustrate the few-shot performance of \ourmethod{} using different number of tasks, and relevant domain baselines (e.g., PostgreSQL, BAO for query optimization). Horizontal dashed lines indicate the performance of various full BO runs and other optimizers, shown for comparison. These results demonstrate that \ourmethod{}'s few-shot performance is often comparable to or surpasses that of full BO runs.} 
\label{fig:scatter-plots}
\end{figure*}

The transformer-based methods demonstrate notable limitations in \cref{fig:scatter-plots,,fig:oracle_calls}. Optformer achieves performance slightly worse than STBO on the database task while showing significantly poorer results on the peptide task. LLAMBO performs substantially worse across both domains, showing minimal progress during optimization. Due to its computational demands—requiring LLM inference for both surrogate modeling and acquisition—LLAMBO completed fewer than 100 optimization steps within our token budget constraints.

We find that on both tasks once \ourmethod{} reaches a sufficient scale, it begins to generate \emph{initialization data} for BO that is significantly better performing than the final results found by all baseline methods, including STBO, DKT/FSBO, the GP expert methods, and transformer-based approaches.

\vspace{-1.5ex}
\paragraph{BOLT as a one- and few-shot optimizer.} In \cref{fig:scatter-plots}, \ourmethod{} demonstrates strong few-shot generalization capabilities, even achieving single-shot performance competitive with traditional approaches. In query optimization, all \ourmethod{} variants outperform the top BAO solution within 5 samples. Notably, \ourmethod{}-1138 and \ourmethod{}-1426 surpass PostgreSQL in a single sample, indicating their potential for rapid deployment in low latency scenarios. The performance of \ourmethod{} consistently improves with more iterations across both tasks, except at 50 samples on the query optimization task, where \ourmethod{}‐1138 slightly outperforms \ourmethod{}‐1426. This may be due to variances in LLM sample generation or training. Overall, results confirm \ourmethod{}'s robustness when scaling to thousands of tasks. We further compare our few‐shot performance (\cref{fig:scatter-plots}, before x-axis break) to full BO runs (\cref{fig:scatter-plots} after x-axis break). In both tasks, \ourmethod{} achieves few shot results comparable to the full BO runs.

\subsection{Ablation Studies}

\begin{figure}[t]
  \begin{minipage}[b]{.48\linewidth}
    \centering
    \includegraphics[width=\linewidth]{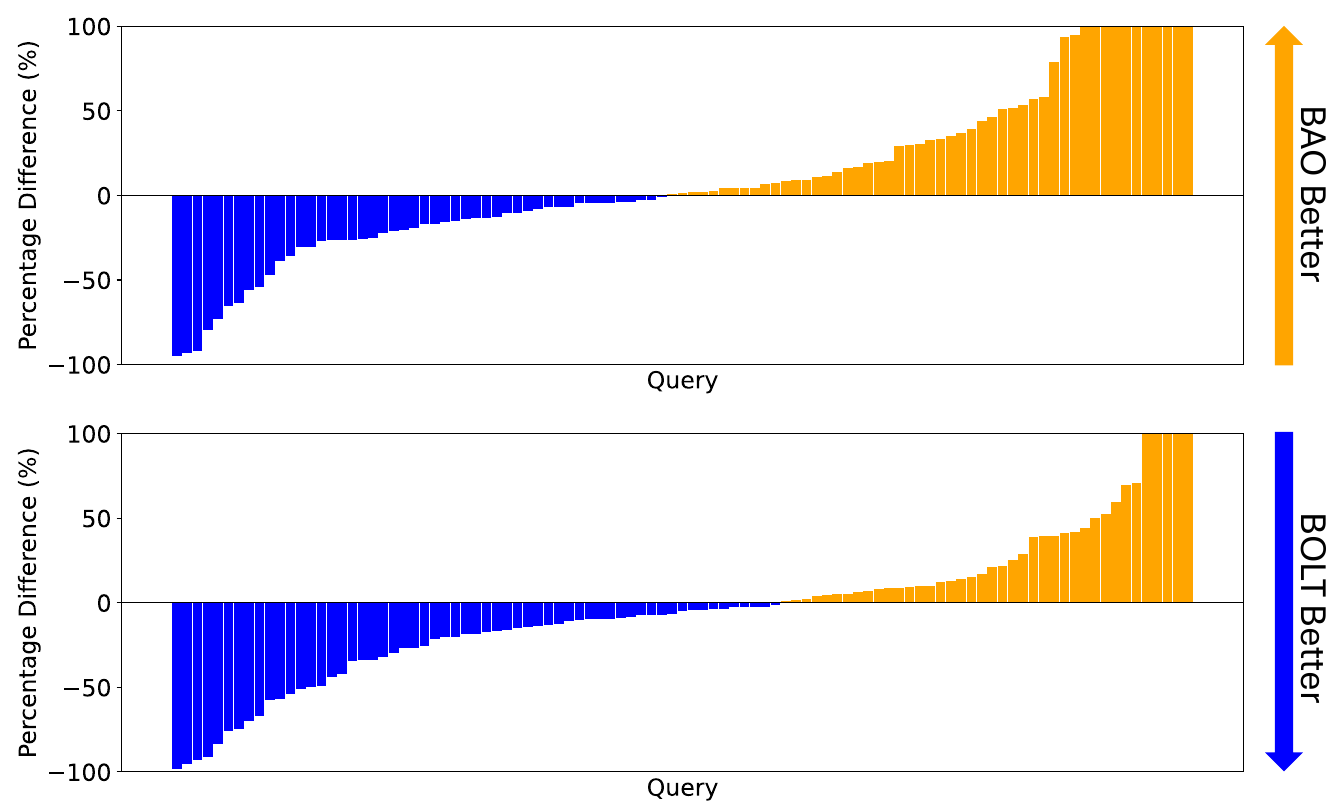}
    \captionof{figure}{Comparing one-shot LLM generated initialization to the best of the 50 BAO plans. \textbf{(Top)}: One-shot generation from \ourmethod{}-$1426$, sampling temperature $T=0.7$. \textbf{(Bottom)}: One-shot generation from \ourmethod{}-$1426$, greedy sampling $T=0.0$. In the one-shot setting, purely greedy sampling is better (lower, more blue).}
    \label{fig:round3-vs-bao}
  \end{minipage}\hfill
  \begin{minipage}[b]{.48\linewidth}
    \centering
    \begin{tabular}{cc}
    \toprule
    \textbf{Method} & \textbf{Best@50} \\
    \midrule
    LLM (\ourmethod{}-50) no SA& 87.84 \\
    LLM (\ourmethod{}-893) no SA& 82.31 \\
    LLM (\ourmethod{}-1138) no SA& 78.16 \\
    LLM (\ourmethod{}-1426) no SA& 63.68 \\
    \midrule 
    LLM (\ourmethod{}-50)  & \textbf{82.25} \\
    LLM (\ourmethod{}-893) & \textbf{63.05} \\
    LLM (\ourmethod{}-1138) & \textbf{61.46} \\
    LLM (\ourmethod{}-1426) & \textbf{61.54} \\
    \bottomrule
    \end{tabular}
    \captionof{table}{Ablation study for self augmentation (SA) conducted on the query optimization task. For each of two LLMs with different training task sizes, we perform SA and generate 50 query plans from the LLM. We measure the best summed query execution time across the validation tasks from among these 50 samples.}
    \label{table:si_helps}
  \end{minipage}
\end{figure}

\vspace{-1.5ex}
\paragraph{LLM self-augmentation.} \label{querysa}
We investigate whether self-augmentation as outlined in \cref{alg:boot} can improve LLM performance while avoiding the computational expense of the inner-loop BO on the query plan optimization task. We apply the self-augmentation process to \ourmethod{}-1138 and \ourmethod{}-1426, generating 10 samples from each across all $2,933$ training tasks, keeping only queries that outperform the best query plan from BAO's solutions. We then use these two dataset as additional fine-tuning to create self-augmented versions of the LLMs.

\Cref{table:si_helps} shows that this self-augmentation yields substantial improvements even without additional tasks optimized by BO. Both self-augmented models converge to a similar performance level, achieving a summed runtime of about 62 seconds across the 100 validation queries. This convergence suggests a natural performance plateau after training on either 1,500 tasks (\ourmethod{}-1426) or 1,100 tasks plus self-generated samples (\ourmethod{}-1138+SA). The consistency of this plateau across different training approaches further demonstrates \ourmethod{}'s robustness when scaling to large task sets.
This self-augmentation experiment indicates that once the LLM has been fine-tuned to sufficient performance, it can generate additional fine-tuning data, reducing the number of BO runs required. Additionally, our framework scales well with more training tasks without performance loss.

\vspace{-1.5ex}
\paragraph{Greedy decoding.} In all other experiments, we use a constant sampling temperature of 0.7 for response generation. In \cref{fig:round3-vs-bao}, we demonstrate that greedy decoding (i.e., setting the temperature to 0) improves one-shot performance. Results in \cref{fig:round3-vs-bao} show that this strategy generates a larger set of queries that surpasses the best among 50 BAO-generated queries by a wider margin.

\begin{wraptable}{r}{0.6\textwidth}
\centering
\begin{tabular}{cccc}
\toprule
\textbf{Best@} & \textbf{\ourmethod{}-1138} & \textbf{\ourmethod{}-1138$^\star$} & \textbf{\ourmethod{}-1426} \\
\midrule
\textbf{Best@50} & 78.16 & 64.03 & 63.68 \\
\textbf{Best@20} & 82.59 & 70.52 & 66.23 \\
\textbf{Best@10} & 90.19 & 74.40 & 70.21 \\
\textbf{Best@5} & 102.99 & 85.21 & 76.28 \\ 
\textbf{Best@2} & 127.97 & 129.26 & 102.29 \\ 
\textbf{Best@1} & 202.04 & 193.64 & 160.22 \\ 
\bottomrule
\end{tabular}
\caption{Comparing LLMs fine-tuned with \textbf{(Left) }data from 1138 tasks, \textbf{(Right)} data from 1426 tasks, and \textbf{(Middle)} data from 1138 tasks, but including the extra data from \ourmethod{}-1426, and removing data from older tasks. 
}
\label{table:better_more_data}
\end{wraptable}

\vspace{-1.5ex}
\paragraph{Impact of data quality on training.} We perform an ablation to assess the importance of using "better" versus "more" training data for fine-tuning LLMs through iterations of \ourmethod{}. Starting with the \ourmethod{}-$1138$ model, we collect top solutions from a new BO round and train two variants: 1) \ourmethod{}-$1426$, which adds all new solutions to the original \ourmethod{}-$1138$ set. 2) \ourmethod{}-$1138^\star$, which instead \textit{replaces} an equal number of \textit{old} solutions to maintain the same training set size. As shown in \cref{table:better_more_data}, both benefit from higher-quality data, suggesting ``better'' data boosts performance. However, \ourmethod{}-$1138^\star$ underperforms \ourmethod{}-$1426$, which incorporates more and better data, confirming that both factors enhance model performance.

\section{Related Work}

\vspace{-1.5ex}
\paragraph{Language models as optimizers.}
Large language models (LLMs) have recently gained attention as sequence optimizers capable of tackling diverse black-box tasks where direct gradient information is unavailable or difficult to compute. LLM-based optimizers leverage the flexibility of natural language prompts to encode candidate solutions, constraints, and relevant task information. Methods like OPRO illustrates how iterative prompting can refine solutions \citep{opro, zelikman2024selftaughtoptimizerstoprecursively}, while other approaches integrate self-improving strategies that reuse high-performing LLM outputs for further fine-tuning \citep{shypula2024code}. This set of techniques has been applied to biophysical domains such as molecular design and protein engineering, where the LLM proposes mutations to enhance certain properties, as well as to program optimization tasks where the LLM speeds up code execution time \citep{shypula2024code, chemsearchllm, Madani2023-sf}.

\vspace{-1.5ex}
\paragraph{Database optimization.}
Recent work has applied Bayesian optimization (BO) to improve overall database performance~\cite{db_tune_survery,dbtune,dbtune_workload_shift} by tuning the parameters of the database configuration. 

As far as we are aware, \citet{sigmod-arxiv} were the first to apply BO to the specific setting of database query plan optimization considered in this paper. 
Other work has applied reinforcement learning (RL) to query plan optimization \citep{neo,balsa,lero}. RL query optimizers learn from mistakes and improve performance over time. Unlike BO, however, RL requires large supervised datasets for pre-training and typically aims to minimize cumulative query latency rather than achieving the lowest possible latency.
\section{Discussion and limitations}
\label{sec:limits}

\vspace{-1.5ex}
We first highlight a few limitations. First, our approach not only requires that all tasks in a problem setting have the same input domain (a problem that has been explored e.g. by \citet{fan2022hyperboplus}). We further require the existence of a task description context $C[f_{t}]$ that can be used in an LLM prompt to define the task. This is likely more difficult, e.g., for hyperparameter optimization, where the primary thing distinguishing tasks is the data that the models are to be trained on; for this setting, approaches such as \cite{wang2024pre} are likely more appropriate. Finally, we note that the cost of LLM fine-tuning is significantly higher than simple gradient updates of a shared feature extractor. 

Despite these limitations, in two real-world applications where \ourmethod{} was applicable it yielded strong results. Few-shot generation matched ``from scratch'' BO runs, and initializing BO from the LLM samples often improved performance further. Moreover, the interplay between the LLM and Bayesian optimization is noteworthy. Despite interest in using LLMs for optimization \citep{opro, zelikman2024selftaughtoptimizerstoprecursively,shypula2024code, chemsearchllm, Madani2023-sf}, finding initial strong solutions to fine-tune them is challenging in some domains. Bayesian optimization, by offering in-depth search, is an excellent candidate for this.

\newpage
\begin{ack}
N. Maus was supported by the National Science Foundation Graduate Research Fellowship; 
J. R. Gardner was supported by NSF awards IIS-2145644 and DBI-2400135;
C. de la Fuente-Nunez was supported by NIH grant R35GM138201, and by Defense Threat Reduction Agency grants HDTRA11810041, HDTRA1-21-1-0014, and HDTRA1-23-1-0001. 
\end{ack}

\bibliographystyle{plainnat}
\bibliography{references}




\newpage
\section*{Appendix}
\appendix

\section{Prompt details.}
Figures \cref{fig:db_train_prompt} and \cref{fig:peptide_train_prompt} illustrate the prompt templates used for generating optimized query plans and peptide sequences, with \textsc{GPT-4o-mini-0718}. Figure 4 shows the template for database query optimization, where the system acts as an assistant providing efficient join orderings for a given SQL query. Figure 5 displays the template for antimicrobial peptide design, where the system's role is to modify peptide sequences to enhance antimicrobial activity.


\begin{figure}[h]
    \centering
    \begin{tcolorbox}[colframe=black, colback=white, width=\textwidth]
        \begin{lstlisting}[breaklines=true, basicstyle=\footnotesize\ttfamily, frame=none]
System: You are a helpful assistant that provides efficient join orderings for given queries.
User: {SQL query to be optimized}
Assistant: {Optimized query plan}
        \end{lstlisting}
    \end{tcolorbox}
    \caption{The prompt template used for prompting \gptfourmini for generating optimized query plans.}
    \label{fig:db_train_prompt}
\end{figure}  
\label{appendix:peptide_prompt}
\begin{figure*}[h]
    \centering
    \begin{tcolorbox}[colframe=black, colback=white, width=\textwidth]
        \begin{lstlisting}[breaklines=true, basicstyle=\footnotesize\ttfamily, frame=none]
System: You are a specialized assistant that modifies peptide sequences to enhance antimicrobial activity. Make up to 25% sequence modifications based on known antimicrobial peptide properties such as: positive charge, hydrophobicity, and amphipathicity.
User: {Seed peptide to be modified}
Assistant: {Modified peptide}
        \end{lstlisting}
    \end{tcolorbox}
    \caption{The prompt template used for prompting \gptfourmini for generating optimized peptide sequences.}
    \label{fig:peptide_train_prompt}
\end{figure*}
\label{appendix:baseline_info}

\section{Compute details.}
\label{sec:compute} 
This section provides details about all computing resources used to run all experiments and create all results provided in this paper. For experiments, we use GPU compute, using a combination of 18 GPUs from our internal compute cluster. The internal cluster has 2 machines with 18 total GPU nodes. One machine has 8 NVIDIA RTX A6000 GPU with 48GB of VRAM, and has 2 sockets with 48 logical threads per socket. The other machine has 10 NVIDIA RTX A5000 GPUs with 24 GB of VRAM, and has 2 sockets with 24 logical threads per socket. Running all experiments needed to produce all results in this work required roughly  $25000$ total GPU hours.

\section{Implementation details.}
\subsection{DKT/FSBO implementation details.}
\label{gpdetail}

For the antimicrobial peptide design task, a \texttt{PPGPR} model was trained using the \texttt{GPyTorch} module. This model employed a fully connected network with two hidden layers, each having a dimension of 256. Training parameters included a batch size of 128, a learning rate of 0.01, and 1024 inducing points for all peptide design experiments.

For the database query plan optimization task, the \texttt{PPGPR} model utilized a fully connected network with two hidden layers, each with a dimension of 64. A batch size of 16, a learning rate of 0.01, and 1024 inducing points were used for these experiments.

For both tasks, 50 STBO optimization trajectories were randomly selected. The DKT/FSBO-$10/20/50$ models were trained using the first 10, 20, or 50 of these trajectories, respectively. All models were trained for 20 epochs.

\subsection{POGPE/SGPE implementation details.}
\label{ensemble_detail}

Similarly to \cref{gpdetail}, for the antimicrobial peptide design task, each expert model was a \texttt{PPGPR} model implemented with \texttt{GPyTorch}, with a fully connected network with two hidden layers, each with a dimension of 256. A batch size of 128, a learning rate of 0.01, and 1024 inducing points were used.

For the database query plan optimization task, each expert model uses a fully connected network with two hidden layers, each with a dimension of 64. The training used a batch size of 16, a learning rate of 0.01, and 1024 inducing points.

The same 50 STBO optimization trajectories from \cref{gpdetail} were used, and the first $5/10/20$ trajectories were used to train the POGPE/SGPE expert models. In POGPE, all experts were weighted equally. For SGPE experiments, the weighting scheme from \citet{pogpe} was adopted, where the independent GP for the target dataset carries the same weight as the entire set of experts.

\subsection{Optformer implementation details.}
\label{optformer_detail}

For both the query plan optimization and antimicrobial peptide design tasks, \textsc{GPT-4o-mini-0718} was fine-tuned on past optimization trajectories. To stay within context window limits, a maximum input context length of 100 trials and an output of 20 trials were used. The objective value ranges for both tasks were discretized into 1000 equidistant points. The training sets were constructed by randomly subsampling two trajectories of length 120 from the optimization trajectories. The query plan optimization task is trained on 27.4 million tokens and the antimicrobial peptide design task is trained on 4.8 million tokens. Both models were trained for 1 epoch with a batch size of 20 and an OpenAI learning rate multiplier of 1.8. 

Optimization was initialized using the same points as single-task BO. During inference, a constant temperature of 0.7 was used. To manage inference token usage, a batch size of 20 was employed, where the model predicted the next 20 trials based on the previous 100 trials. This was important as experiments ran for 4,000 (query plan) or 20,000 (peptide design) trials.

\subsection{LLAMBO implementation details.}
\label{llambo_detail}

The end-to-end LLAMBO method was utilized, leveraging \textsc{GPT-4o-mini-0718} for several components: generating candidate solutions, serving as a surrogate model for the objective function (via in-context learning), and acting as a conditional sampler to generate candidates for specific target values. Similar to Optformer, a maximum input context window of 100 trials was enforced to prevent exceeding context limits.

The hyperparameters from the original LLAMBO paper were adopted, including an exploration hyperparameter $\alpha=0.1$, and $M=20$. For the surrogate model, we sample $K=10$ MC predictions to compute the empirical estimates. Consistent with the LLAMBO paper, we use the same sampling parameters with a temperature of $0.7$ and top\_p of $0.95$. A limit of 10 million maximum input tokens per experiment was used to manage computational costs.

\subsection{LLM self-augmentation details.}

For the antimicrobial peptide design task, 200 samples were generated for each of the initial $800$ training peptides during each self-augmentation round. Any peptides with a predicted MIC below 8 (indicating significant antimicrobial activity) were added to the training set for the subsequent round of \ourmethod{}.

For the database query plan optimization task, 10 samples were generated for each of the 2,933 training queries. Query plans with a runtime lower than the best plan generated by the BAO optimizer were added to the training set for the next round of \ourmethod{}.

\section{Additional abalations.}
\subsection{Open source LLMs.} 

To explore the viability of open-source models for \ourmethod{}, \textsc{QWEN-2.5-7B} and \textsc{LLAMA-3.1-8B} were fine-tuned using the identical dataset that created \ourmethod{}-1426 from \textsc{GPT-4o-mini-0718} for the database query optimization task. For evaluation, 50 query plans were generated from each LLM using a sampling temperature of 0.7. The best summed query execution time across the validation tasks from these 50 samples was compared.

Both models were fine-tuned on 4 NVIDIA RTX A6000 GPUs using a per-device batch size of 4, a learning rate of 1e-5 with the AdamW optimizer, and 5 training epochs. The results, shown in \cref{table:open_source_llms_db}, indicate that \textsc{QWEN-2.5-7B} performed slightly worse than fine-tuned \textsc{GPT-4o-mini-0718}, while \textsc{LLAMA-3.1-8B} showed significantly lower performance. Due to the extensive number of inference calls and multiple fine-tuning rounds required by BOLT, the primary experiments were conducted using the OpenAI API due to hardware resource limitations.

\begin{table}[htbp]
\centering
\begin{tabular}{lc}
\toprule
\textbf{Model} & \textbf{Summed runtime} \\
\midrule
\texttt{GPT-4o-mini-0718} & 61.54  \\
\texttt{QWEN-2.5-7B} & 62.04  \\
\texttt{LLAMA-3.1-8B} & 155.55  \\
\bottomrule
\end{tabular}
\vspace{1.0pt}
\caption{Comparing open source LLMs fine-tuned with data used to fine-tune \ourmethod{}-1426 against OpenAI models fine-tuned on the same data.}
\label{table:open_source_llms_db}
\end{table}

\subsection{Initializing with previous solutions.}

We investigate whether utilizing the best solutions from previously encountered problems provides a sufficiently strong starting point, potentially removing the need for \ourmethod{}'s specialized initialization. To assess this, we collect the best-performing solution from each task previously optimized by \ourmethod{}-1426. This collection of "previous solutions" is then evaluated on the first 10 validation tasks. Their aggregated performance (summed runtime) is compared against the initialization provided by \ourmethod{} (specifically, \ourmethod{}-1426, \ourmethod{}-1138, and \ourmethod{}-893), a standard Bayesian Optimization baseline (STBO), and \ourmethod{}-1426 when further refined by a full BO run (\ourmethod{}-1426 + BO). The comparison shows that initializing with "previous solutions" yields a higher summed runtime than any of the \ourmethod{} initializations or the STBO baseline, suggesting that simply reusing prior optimal solutions is less effective than \ourmethod{}'s approach. The results are presented in Table~\ref{table:init_with_prev}.

\begin{table}[htbp]
\centering
\begin{tabular}{lc}
\toprule
\textbf{Method} & \textbf{Summed runtime} \\
\midrule
\ourmethod{}-1426 + BO & 6.43 \\
\ourmethod{}-1426 & 7.21  \\
\ourmethod{}-1138 & 8.65  \\
\ourmethod{}-893 & 9.08  \\
\texttt{STBO} & 8.23  \\
previous solutions & 9.18  \\
\bottomrule
\end{tabular}
\vspace{1.0pt}
\caption{Comparison of summed runtimes on the first 10 validation tasks when initializing with different strategies. "\ourmethod{}-1426 + BO" represents full Bayesian optimization initialized by \ourmethod{}-1426. "\ourmethod{}-1426", "\ourmethod{}-1138", and "\ourmethod{}-893" show the performance of \ourmethod{}'s initialization alone. "STBO" is a standard Bayesian optimization baseline. "Previous solutions" refers to using the best known solutions from all tasks previously seen by \ourmethod{}-1426 as the initialization set.}
\label{table:init_with_prev}
\end{table}

\newpage

\section{Broader Impacts.}
\label{sec:broaderimpact}

This research introduces \ourmethod{}, a method leveraging large language models to enhance multi-task Bayesian optimization, with demonstrated applications in antimicrobial peptide design and database query optimization. The potential to accelerate the discovery of novel peptides could significantly benefit public health, particularly in combating antimicrobial resistance. Similarly, improving database query efficiency can lead to substantial computational and energy savings across many industries.

However, we acknowledge the potential for AI misuse in biological design. In applying these powerful methods, expert oversight, rigorous validation, and adherence to established safety and regulatory frameworks must be highlighted. Additionally, the use of large-scale LLMs raises considerations regarding computational accessibility and responsible AI development.

\end{document}